\documentclass{article}

\usepackage{microtype}
\usepackage{graphicx}
\usepackage{subfigure}
\usepackage{booktabs} 

\usepackage{hyperref}

\usepackage[accepted]{mlsys2023}

\usepackage{multicol}
\usepackage{float}

\usepackage{fancyhdr}
\usepackage{paralist}
\renewenvironment{itemize}[1]{\begin{compactitem}#1}{\end{compactitem}}
\renewenvironment{enumerate}[1]{\begin{compactenum}#1}{\end{compactenum}}

\usepackage{amsmath}
\DeclareMathOperator*{\argmax}{arg\,max}

\usepackage{amsfonts}

\mlsystitlerunning{X-RLflow: Graph Reinforcement Learning for Neural Network Subgraphs Transformation}

\begin{document}

\twocolumn[
\mlsystitle{X-RLflow: Graph Reinforcement Learning for Neural Network Subgraphs Transformation}



\begin{mlsysauthorlist}
\mlsysauthor{Guoliang He}{cam}
\mlsysauthor{Sean Parker }{cam}
\mlsysauthor{Eiko Yoneki}{cam}
\end{mlsysauthorlist}

\mlsysaffiliation{cam}{University of Cambridge}

\mlsyscorrespondingauthor{Eiko Yoneki}{eiko.yoneki@cl.cam.ac.uk}

\mlsyskeywords{Reinforcement Learning, Tensor Graph Superoptimisation}

\vskip 0.3in

\begin{abstract}

Tensor graph superoptimisation systems perform a sequence of subgraph substitution to neural networks, to find the optimal computation graph structure. Such a graph transformation process naturally falls into the framework of sequential decision-making, and existing systems typically employ a greedy search approach, which cannot explore the whole search space as it cannot tolerate a temporary loss of performance. In this paper, we address the tensor graph superoptimisation problem by exploring an alternative search approach, reinforcement learning (RL). Our proposed approach, \textbf{X-RLflow}, can learn to perform neural network dataflow graph rewriting, which substitutes a subgraph one at a time. \textbf{X-RLflow} is based on a model-free RL agent that uses a graph neural network (GNN) to encode the target computation graph and outputs a transformed computation graph iteratively. We show that our approach can outperform state-of-the-art superoptimisation systems over a range of deep learning models and achieve by up to $40\%$ on those that are based on transformer-style architectures.

\end{abstract}
]



\printAffiliationsAndNotice{}  

\section{Introduction}
Recent modern software has key components that are underpinned by machine learning models, specifically, deep neural networks (DNNs). Over the past decade, there has been a focus on developing frameworks that provide tools using which we can design, train and evaluate these deep learning models.

A common internal representation for neural networks inside deep learning frameworks is that of a computation graph; a directed acyclic graph where nodes represent a specific computation and edges the paths where data is transferred. With the graph representation, frameworks such as TensorFlow \cite{tensorflow2015whitepaper, 199317} and PyTorch \cite{paszke2019pytorch} apply optimisations to reduce computation resources during inference.

Currently, the majority of graph-level optimisations in deep learning frameworks are performed using manually defined heuristics. For example. TensorFlow, TensorRT \cite{tensorrt2017}, and TVM \cite{chen2018tvm} perform fusion to a computation graph by using rule-based strategies. While such heuristics apply to current architectures, as network design is consistently evolving, new rules are being constantly discovered and managing heuristics quickly becomes unwieldy. 

To mitigate this problem, recent work, namely TASO \cite{jia2019taso}, has shown an automatic cost-based search can replace heuristics to perform tensor graph rewriting. TASO first generates a set of rewrite rules by enumerating operators and then applies the generated rewrite rules to tensor programs via a backtracking search. However, such a backtracking search approach may not fully explore the potential search space due to the lack of planning in cost-based optimisation. As a step towards resolving the planning issue, this work explores the use of reinforcement learning (RL). RL is an area of machine learning in which an agent learns to act optimally, given a state and a suitable reward function, through interactions with an environment.

In this paper, we introduce \textbf{X-RLflow}, a tensor graph superoptimiser that uses Reinforcement Learning (RL) for automating tensor graph transformation. RL is known to be a better search methodology than a backtracking search for its planning for long-term reward, and thus it is more likely to discover the globally optimal tensor graph structure. 

Applying RL to the tensor graph superoptimisation domain requires non-trivial modification because it takes the target computation graph as input and outputs a better candidate that is transformed by rewrite rules. As multiple rewrite rules are applicable to every iteration, choosing the best one is difficult because it also determines how the tensor graph can be transformed in subsequent iterations. As a result, the agent must learn to act optimally at each iteration, making the decision that is not only good for the current iteration but also good for the long term. Specifically, the current computation graph and all potential substitution candidates are encoded via a graph neural network (GNN) to a representation and the graph representation is later fed into a policy network and a value network respectively to produce the action probability and value estimate, as in the standard actor-critic framework \cite{ac}. This process is performed iteratively until no rewrite rules are applied or the agent outputs a No-Op action, which then terminates the tensor graph superoptimisation process. The final optimised graph runs end-to-end inference to measure its inference latency. 

The use of RL for tensor graph superoptimisation also enables using end-to-end inference latency as the feedback signal. We find that the results from cost modelling can deviate from the end-to-end inference latency by up to $24\%$, and thus using cost modelling as the feedback signal to guide the transformation process may lead to sub-optimal results. On the other hand, running end-to-end inference causes significant measurement overhead, but the overhead can be amortised by evaluating sparsely. RL by design can work well in a sparse or delayed reward scenario, and it can still learn to maximise the long-term reward.

We also find that X-RLflow offers the ability to generalise. This is because the policy of RL is parametrised by a neural network, and it can be reused for performing inference once trained. We show that X-RLflow can generalise to various tensor shapes after it is trained in a static tensor shape environment. X-RLflow is available as open source \footnote{\url{https://github.com/ucamrl/xrlflow.git}}.

To summarise, our contributions are:

\begin{itemize}


\item We design an RL agent and environment for automatically selecting a sequence of subgraph transformation. 
\vskip 1mm

\item Our approach works well for a wide range of deep learning models and is especially powerful for transformer-style architectures demonstrated by outperforming state-of-the-art methods by up to $40\%$.
\vskip 1mm

\item We provide a detailed discussion and analysis of our solution as well as a comparison to the state-of-the-art methods in published literature.
\vskip 1mm

\item This work, to the best of our knowledge, is the first that has applied reinforcement learning in the tensor graph structure superoptimisation domain.

\end{itemize}

\section{Background and Motivation}

\subsection{Computation graphs for neural networks}

\begin{figure}
  \begin{center}
  \includegraphics[width=0.75\columnwidth]{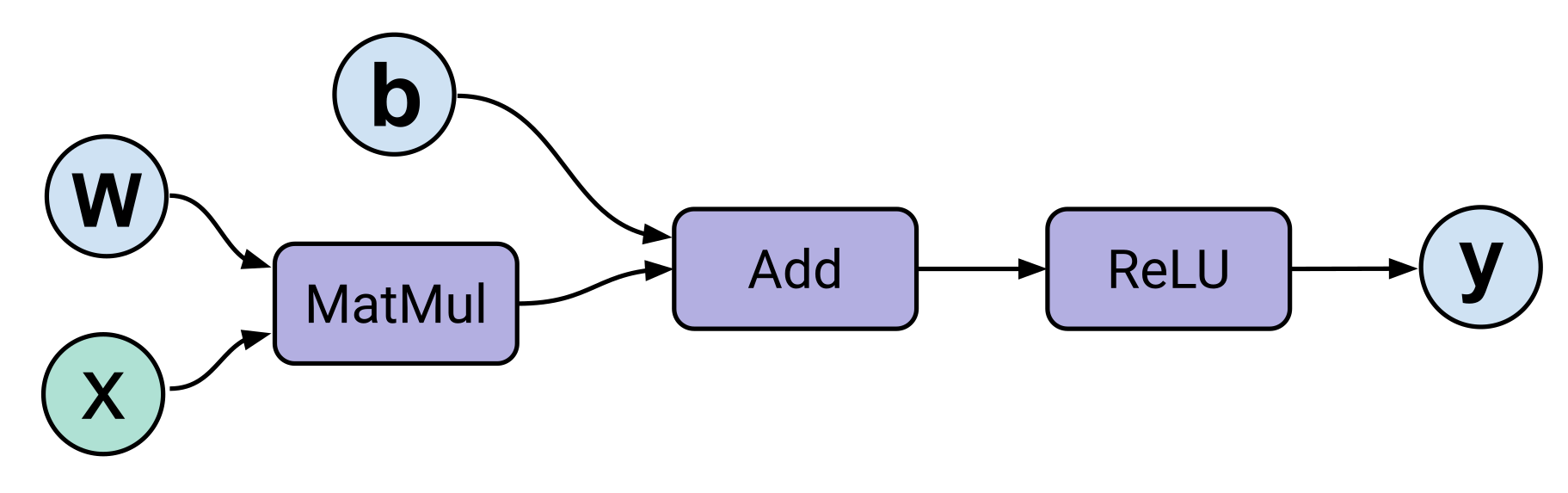}
  \end{center}
  \caption[Single perceptron as a computation graph]{The graphical representation of a dense linear layer, where tensor operators are nodes and the directed edges show the flow of tensors through the graph.}
  \label{fig:bg:perceptron}
\end{figure}

To enable performance optimisation for DNNs, deep learning frameworks and compilers represent DNNs as dataflow/computation/tensor graphs, where tensor operations become nodes, and tensors are edges. Figure \ref{fig:bg:perceptron} shows how a dense linear layer, $y = \texttt{ReLU}(\mathbf{w} \cdot \mathbf{x} + b)$, can be represented as a computation graph. The terms dataflow graph, computation graph and tensor graph are used interchangeably in deep learning frameworks.

\subsection{Tensor graph structure superoptimisation systems}
\label{sec::Tensor graph structure superoptimisation systems}

With the graphical intermediate representation (IR), tensor graph superoptimisation systems attempt to perform subgraph substitution, aiming to reduce the end-to-end inference latency of target DNNs. 

\subsubsection{Existing systems}

TASO \cite{jia2019taso} is the first system dedicated to tensor graph structure superoptimisation. It first generates a set of rewrite rules by enumerating a list of pre-defined operators and then performs optimisation by substituting the subgraph of the target DNN. During the optimisation phase, TASO uses a cost model to rank all candidates and greedily chooses the best candidate to proceed to the next iteration.

PET \cite{wang2021pet} further builds on TASO and explores partially equivalent subgraph transformation. The rewrite rules developed in TASO only consider fully equivalent subgraph substitution, and PET relaxes this assumption by allowing non-equivalent substitution. Correction kernels are automatically generated to ensure end-to-end equivalency after the substitution.

Tensat \cite{tensat} uses the same superoptimisation systems as TASO, but replaces TASO's optimiser with an equality saturation \cite{equality_saturation} based optimiser. It uses the E-graph data structure to represent many potential graph IRs simultaneously and then extracts the optimal IR from the E-graph.

\subsubsection{Limitations}
\label{sec::limitation}

Existing systems have employed different approaches to find the optimal tensor graph structure, but each comes with its limitations. 

TASO's substitution engine finds the best candidate at each iteration of the transformation, but as the globally optimal tensor graph structure may not be the best option at each iteration, TASO is likely to miss out on the globally optimal tensor graph. Moreover, the performance of candidates is evaluated via the cost model, which assumes the summation of individual operator runtime is the same as the end-to-end inference latency. However as shown in Table \ref{tab: taso cost model vs end2end}, the discrepancy between the measurement of the cost model and the actual end-to-end inference latency can be up to $24\%$. This indicates the best candidate ranked by the cost model may not be the best choice for actual deployment. 

\begin{table}[t]
\caption{Discrepancy between TASO's cost model estimates and TASO's end-to-end inference latency on some unoptimised DNNs. E2E stands for end-to-end inference latency. Time is measured in milliseconds.} 
\label{tab: taso cost model vs end2end}
\vskip 0.15in
\begin{center}
\begin{small}
\begin{sc}
\begin{tabular}{lcccr}
\toprule
DNNs & Cost model & E2E & Diff (\%) \\
\midrule
         DALL-E & 1.8269  & 1.7324 & 5.2\%\\  
         InceptionV3 & 8.3650 & 9.2098 & 10.1\%   \\
         BERT & 1.0453 & 1.1264 & 7.8\%  \\
         Squeezenet & 1.3082 & 1.4006 & 7.1\%  \\
         ResNext-50 & 6.1545 & 7.6498 & 24\%  \\
         T-T & 2.4828 & 2.7281 & 9.9\%  \\
\bottomrule
\end{tabular}
\end{sc}
\end{small}
\end{center}
\vskip -0.1in
\end{table}

PET performs substitution by considering partially equivalent transformation. However, it adopts the same cost modelling principle as TASO, and therefore it suffers from the same cost modelling problem. PET does not provide an end-to-end inference interface in its artefact evaluation, so it is impossible to measure the discrepancy in this case. We also find that PET ignores all element-wise operators' runtime, which may exacerbate the cost modelling problem. 

We further compare PET and TASO on two similar DNNs, but their performances are very different. As shown in Table \ref{tab: taso vs pet}, PET outperforms TASO in ResNet-18 but falls short in ResNext-50. We hypothesise this result is because PET's partially equivalent transformation is very sensitive to the shape of operators. Choosing the right operator shapes may bring significant improvement to partially equivalent transformation, and PET's paper also mentions a larger batch size offers more optimisation opportunities. However, understanding when partially equivalent transformation performs well is beyond the scope of this paper, and as a result, we will focus on TASO in this work.

\begin{table}[H]
    \caption{Comparison of the optimised graph inference latency between PET and TASO in ResNet-18 and ResNext-50. Time is measured in milliseconds.}
    \label{tab: taso vs pet}
\vskip 0.1in
    \centering
    \resizebox{.3\textwidth}{!}{%
    \begin{tabular}{c|c|c}
        & ResNet-18 & ResNext-50  \\ \hline
         PET & 1.9619  & 10.6694  \\  
         TASO & 2.5534 & 6.6453    \\
    \end{tabular}
    }
\end{table}

Tensat employs equality saturation \cite{es}, which leverages the E-graph data structure to compactly represent many potential tensor graphs at the same time. Although in theory, a saturated E-graph can represent all possible IRs, in reality, the E-graph is never saturated due to several reasons. First, a saturated E-graph can be too large to fit into memory, and therefore the E-graph is usually upper-limited by $10000$ nodes. Second, a large E-graph takes a long time to extract the optimal IR, and extracting IRs from a very large E-graph is non-trivial. As a result, Tensat's E-graph is not saturated and it cannot guarantee that its optimised tensor graph structure is optimal.

\subsection{Reinforcement learning}

\subsubsection{Reinforcement learning basics}

We propose to use Reinforcement learning (RL) to tackle the tensor graph structure superoptimisation problem. RL aims to compute a control policy such that an agent can maximise its cumulative reward from the environment. The agent will learn to discover the optimal strategy via a single reward signal.

Formally, RL is a class of learning problems that can be framed as a Markov decision process (MDP) \cite{bellman1957}; they are represented as a 5-tuple $\langle \mathcal{S}, \mathcal{A}, \mathcal{P}_a, \mathcal{R}_a, \rho_0 \rangle$ where:

\begin{itemize}
  \item $\mathcal{S}$, is a finite set of valid states
  \item $\mathcal{A}$, is a finite set of valid actions
  \item $\mathcal{P}_a$, is the transition probability function that an action $a$ in state $s_t$ leads to a state $s'_{t+1}$
  \item $\mathcal{R}_a$, is the reward function, it returns the reward from the environment after taking an action $a$ between state $s_t$ and $s'_{t+1}$
  \item $\rho_0$, is the starting state distribution
\end{itemize}

We aim to compute a policy, denoted by $\pi$, that when given a state $s \in \mathcal{S}$, returns an action $a \in \mathcal{A}$ with the optimisation objective being to find a control policy $\pi^*$ that maximises the \textit{expected reward} from the environment as Equation (\ref{expectedreward}). 

    \vspace{-5mm}
\begin{equation}
  \pi^* = \argmax_\pi \mathbb{E} [ \sum^\infty_{t=0} \gamma^t \mathcal{R}_t ]
  \label{expectedreward}
    \vspace{-3mm}
\end{equation}

Classic RL problems are formulated as MDPs in which we have a finite state space. However, such methods quickly become inefficient with large state spaces for applications such as Atari \cite{mnih2013playing, kaiser2020modelbased} and Go \cite{silver2017mastering}. Therefore, we take advantage of modern deep learning function approximators, such as neural networks, which makes learning the solutions far more efficient in practice. We have seen many successful applications in a wide range of fields, for example, robotic control tasks \cite{openai2019solving}, data centre power management, device placement \cite{addanki2019placeto, mirhoseini2018hierarchical}, and playing both perfect and imperfect information games to a super-human level \cite{silver2016mastering,silver2017mastering}. 

\subsubsection{Graph reinforcement learning}

Applying RL to the tensor graph superoptimisation domain cannot be accomplished by classic RL algorithms. This is because computation graphs are naturally graph-like data and have relationships that cannot be expressed in Euclidean space. Fortunately, graph neural networks (GNNs) is proposed to address learning for graph data, and we provide a high-level overview of commonly used GNNs.

Graph Convolutional Networks (GCNs) \cite{gcn} is one of the simplest yet powerful GNNs, which updates each node's features by message passing with its neighbouring nodes' features. While GCN is effective at learning graph data, it assumes equal weighting of neighbouring nodes. Graph Attentional Networks (GATs) \cite{gat} is later proposed to address this problem, by learning to assign weights to neighbour nodes. As a result, GAT is more expressive and is adopted in this work as a component of the system. 

\subsection{Motivation for RL}
\label{sec. Motivation for RL}

Our motivation for using RL in the tensor graph superoptimisation domain comes from several aspects.

First, RL can tolerate short-term performance decrease to maximise long-term rewards. This is important for tensor graph superoptimisation because the optimised graph is obtained by applying a sequence of subgraph substitution, and the globally optimal tensor graph structure may not be the best candidate at every iteration. As a greedy search engine only considers the best candidate, it is likely to miss out on the globally optimal solution. On the contrary, it has been shown in literature \cite{mnih2013playing, kaiser2020modelbased}, \cite{silver2017mastering} that RL can learn to tolerate short-term loss and maximise the long-term episodic reward.

Second, we want to bypass the cost modelling issue and only use the end-to-end inference latency as the feedback signal to choose among potential candidates. We believe this is necessary for machine learning systems since there are multiple IRs for DNNs during progressive lowering and there is optimisation being done for each IR layer. As a result, optimisation that takes place at a higher layer IR is not aware of the optimisation in lower layer IRs. Therefore, cost modelling at a specific layer is not sufficient to form a good feedback signal.

Using end-to-end inference latency is challenging to equality saturation-based methods because extracting from the E-graph needs per-node cost modelling, and for greedy search-based methods, it brings significant overhead to rank all candidates at each iteration. We find RL a good methodology because by design it can work in a sparse or delayed reward scenario. In our design, we perform an end-to-end inference every $N$ iteration, where $N$ is a hyper-parameter and it controls the trade-off between the frequency of reward signals and the measurement overhead. We argue that the accuracy and measurement overhead trade-off widely exists in performance optimisation and the use of RL enables the controlling of this trade-off.

Finally, RL offers a generalisation ability. Deep learning systems often assume statically known tensor shapes and tensor graphs for optimisation, and if any of the two changes, the optimisation must start from scratch. However, there is a practical need of changing the tensor shapes. For example, a language model may need to be compiled multiple times independently because its input text length may vary and so is its corresponding tensor shapes. By applying RL, it should be able to generalise to various tensor shapes. This is because by varying the tensor shape, the tensor graph structure is still the same. Therefore, we only need to train the RL agent once and let it generalise to various tensor shapes of the target DNN.

There is recent work on model-based RL that offers better generalisation ability. For example, the world model \cite{wm} learns the dynamics of the environment, such that the agent can be trained in a latent space. The main benefit of the world model is sample efficiency because the agent does not need to interact with the actual environment. Moreover, the world model enables planning before making decisions. Unfortunately, learning a world model is difficult and an imperfect world model may deviate from the actual environment too much, such that even if the agent learns to act optimally in the world model, it fails to achieve good performance in reality. As such, we focus on model-free agents in this work.

GO \cite{go-1} is a model-free agent that tries to generalise RL to unseen tenor graph structures. To do this, they train RL agents on multiple graphs and evaluate them on held-out graphs. This is much more computationally demanding, so we only focus on generalisation to tensor shapes in this work.

\section{X-RLflow}
\label{sec::optnns}

In this section, we introduce X-RLflow, a tensor graph superoptimiser that uses RL for automatic subgraph substitutions. X-RLflow encapsulates the tensor graph transformation process as the environment transition, and thus RL can iteratively transform the computation graph. We provide a detailed description of each component of X-RLflow in this section. 


\subsection{Computation graphs}

We use the same computation graph representation as in TASO. That is, users can manually define the computation graph via TASO's programming interface, or load pre-trained models to the system. The pre-trained models from existing tensor frameworks, such as TensorFlow \cite{tensorflow2015whitepaper, 199317}, PyTorch \cite{paszke2019pytorch} and MXNet \cite{chen2015mxnet}, can be converted to a unified ONNX format \cite{bai2019onnx}, which then can be parsed to TASO's computation graph representation. After the superoptimisation, users can export the optimised graph back to the ONNX format, and deploy it to different backends.

\subsection{Graph rewrite rules}

\begin{figure}[ht]
  \centering
  \includegraphics[width=0.8\columnwidth]{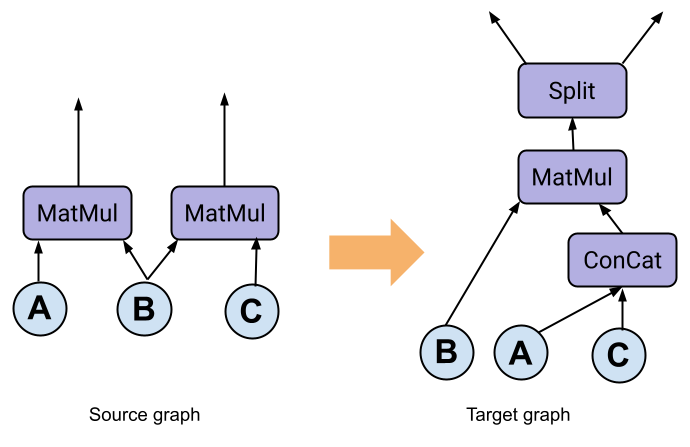}
    \vspace*{-3mm}
  \caption{Example of a TASO's rewrite rule. Applying this rule means performing pattern matching to the target computation graph and substituting the source graph with the target graph. Note that there may be multiple matches in the target computation graph, in which case applying these rules generates multiple transformed candidates.}
  \label{fig: rewrite-rule}
\end{figure}

The tensor graph rewrite rules are generated by TASO's generator. Those rules are generated before the optimisation phase, by enumerating a list of primitive operators up to a constant, and they are serialised to a text file. At the beginning of the optimisation phase, rewrite rules are de-serialised from the text file and activated. There are in total $150$ rewrite rules, and one example of a graph rewrite rule is shown as per Figure \ref{fig: rewrite-rule}.

Given a target DNN, a candidate will be generated by pattern-matching a rewrite rule to the target computation graph. At each iteration of the optimisation, there are typically multiple matching rules to multiple locations of the target computation graph, and therefore multiple candidates are generated. Those candidates are put into a cache, and X-RLflow selects one candidate as the transformation applied to the target DNN. More details about the selection process are provided in the next section.

\subsection{Reinforcement learning formulation}

\subsubsection{System environment}
\label{sec:prob:subsec:sysenv}

We encapsulate the tensor graph transformation process as the environment transition in the standard RL formulation. There exists an open-source environment implementation that provides standardised APIs for RL agents, such as the OpenAI Gym \cite{brockman2016openai}, in which users can extend and write their environment transition logic. In this work, we design an environment that follows the OpenAI Gym API standard stepping an environment, that is, we have a \texttt{step()} function that hides the complexity of the graph transformation process and exposes a unified API to RL agents, and a \texttt{reset()} function that resets the transformation process back to the initial stage. Those are essential functions for the environmental transition. Note that this environment is re-usable for diverse and future RL algorithms.

\begin{figure}[ht]
  \centering
  \includegraphics[width=\columnwidth]{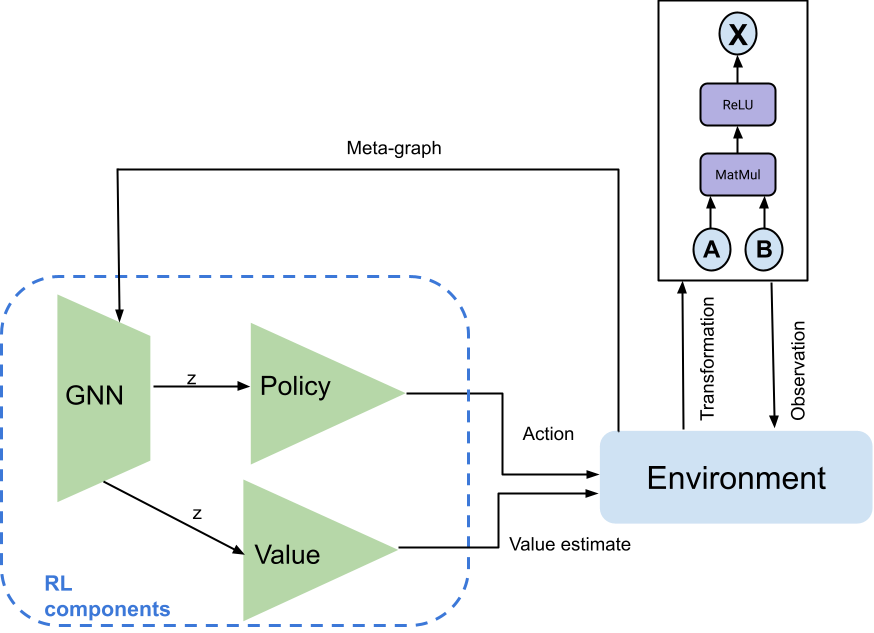}
    \vspace*{-3mm}
  \caption{Interaction between components of X-RLflow.}
  \label{fig:problem:sys-env}
\end{figure}

We made use of the work by Jia et al. \cite{jia2019taso} who provide an open-source version of TASO as part of the backend of the environment, where the rewrite rules pattern matching and tensor graph transformation takes place. At each iteration of the environmental transition, the environment takes as input a computation graph and applies TASO's rewrite rules to all possible locations of the graph via pattern matching, which generates a list of potential candidates. Those potential candidates become the observation for the RL agent.

\subsubsection{State-Action space}
\label{sec:prob:subsec:sap}

The environment generates a list of transformed candidates at each iteration of the optimisation, and the agent makes an action, which is mapped to the index of a potential candidate, indicating the choice of the agent. The selected candidate becomes the graph for the next iteration. This process is repeated until there are no more available candidates, or the agent outputs a No-Op to indicate an earlier termination. The termination indicates the end of the tensor graph transformation process.

To enable decision-making at each iteration, the computation graph and its potential candidates have to be embedded as a state vector, which is done via a graph neural network (GNN). Specifically, to create a graph input, we traverse the target graphs and build node attributes and edge attributes sequentially. For node, we maintain a table of operators and one-hot encode the index of the operator as node attributes. There are around $40$ different tensor operators in total. For edges, we use the corresponding tensor shape as its attributes. For example, the edge attribute $[1, 3, 256, 256]$ means its corresponding tensor has shape $[1, 3, 256, 256]$. For tensors whose rank is less than $4$, zeros are padded to leading dimensions. To stabilise training, we normalise edge attributes via a constant $M$, whose value is detailed in Appendix \ref{appendix:hyper}. We also initialise the global attribute to $0$ for all tensor graphs and update it through a learnable GNN layer. After we build the graph input of the current computation graph and its potential candidates, we batch them into a meta-graph, which is a comprehensive representation of the current state.

One challenge for X-RLflow is the number of potential candidates changes for every iteration. This is because as the computation graph is transformed iteratively and becomes more optimised, there are fewer applicable rewrite rules and therefore fewer potential candidates are generated. As a result, the action space of X-RLflow shrinks during the transformation process. To overcome this issue, we pad the action space to a large constant and use a mask vector to indicate the actual candidates. Specifically, for each meta-graph, we generate an associated boolean mask vector. When the agent outputs the action probability vector, the invalid action is masked out by the boolean vector. This is known as invalid action masking \cite{invalid_action_masking}, which has been studied recently. It effectively turns the gradients to zero if they correspond to an invalid action, and thus resolves the changing action space issue. An alternative way is to penalise the agent when it makes an invalid action and terminates the episode. We find it slows down the training and thus we employ the invalid action masking method.

Figure \ref{fig:problem:sys-env} shows the architecture overview of X-RLflow. The GNN is used to encode the meta-graph input and produce a state representation $z$, and then, the state vector is fed into a policy head and a value head, which output a categorical distribution and a value estimate respectively. The policy head and value head are both two layers of multi-layer perceptrons (MLPs). The action sampled from the categorical distribution is mapped to a candidate index, and the value estimate is stored in a cache for computing the generalised advantage estimate when updating the agent.

The GNN is critical in our system because it enables applied RL to the tensor graph superoptimisation domain and therefore its details will be further described in section \ref{sec:prob:subsec:gnn}.

\subsubsection{Reward function}
\label{sec:prob:subsec:rwd}

Arguably the reward function is regarded as the most important part when designing RL based optimiser. Agents will maximise their policy to get the maximum cumulative reward of episodes, therefore our reward function should encourage the agent to find the optimal tensor graph structures.

Since the optimal tensor graph structure is evaluated via its end-to-end inference latency, the reward function should reward the agent positively when an action decreases the inference latency. To enable flexibility, the environment has an interface where users can register their callback function to compute the reward, and we also have implemented a default reward function that works well in our empirical evaluation.

    \vspace{-5mm}
\begin{equation}
    r_t = \frac{RT_{t-1} - RT_t}{RT_0} * 100
    \label{eq. reward func}
    \vspace{-3mm}
\end{equation}

The default reward function firstly computes the inference time difference between the last inference latency $RT_{t-1}$ and the current inference latency $RT_t$ and then normalises the difference with initial inference time $RT_0$ of the DNN. Intuitively, such a design encourages the RL agent to discover candidates that reduce end-to-end inference latency compared to the current computation graph. The normalisation stabilises the training because the reward is calculated as a percentage speedup and it will not introduce very large positive or negative rewards. We find the default reward function leads to good performance in practice. For invalid action, we simply mask out its probability by assigning a large negative number in the categorical distribution. 

Note that end-to-end inference is run every $N$ iteration for the reason mentioned in section \ref{sec. Motivation for RL}, meaning Equation \ref{eq. reward func} is used every $N$ iteration. When the feedback signal is not available, we simply use a small constant of $0.1$ to reward the agent for continuous exploration. We find that this constant works well in our empirical evaluation, and it can be customised by users via registering a callback function.

\subsubsection{Learning algorithm}
\label{sec:prob:subsec:learning-algorithm}

All learnable components in X-RLflow are trained in an end-to-end fashion. To this end, we adopt the clip variant from the PPO algorithm \cite{ppo}. PPO is an on-policy RL algorithm, meaning it first performs roll-outs for several episodes and uses the collected data to perform an update to its networks. This process is repeated until the number of update round reaches a pre-defined constant.

Being a PPO RL agent, there are multiple available variants of the objective function. The clip objective function chosen for X-RLflow is:

\begin{equation}
    \mathcal{L}_{clip} = -\mathbb{E}_{G} \{ \min(\frac{\pi_\theta}{\pi_{\theta_k}} \cdot A^{\pi_{\theta_k}}, clip(\frac{\pi_\theta}{\pi_{\theta_k}}, 1-\epsilon, 1+\epsilon )A^{\pi_{\theta_k}}) \}
\end{equation}

Where $\pi_\theta$ is the current policy and $\pi_{\theta_k}$ is the old policy. $A^{\pi_{\theta_k}}$ represents the generalised advantages computed given the samples generated from the old policy $\pi_{\theta_k}$, as in \cite{gae}. The clip objective essentially prevents the policy network from being updated too much away from its previous weights. To update the value estimates, we simply compute the mean-square-error of the output from the value head with the target value:

\begin{equation}
    \mathcal{L}_{vf} = \mathbb{E}_{G} \{(V_{\theta}(s_t) -  V_{target})^2\}
\end{equation}

The final objective is the summation of the two losses functions and an entropy term:

\begin{equation}
    J = \mathcal{L}_{clip} + c_1\mathcal{L}_{vf} + c_2 \mathcal{L}_{entropy} 
\end{equation}

Where $c_1$ and $c_2$ are two small constants to weigh the value loss and the entropy loss respectively. The entropy term is calculated from the action probability $\pi_{\theta_k}$ and is meant to constrain the policy network update further.

This particular RL algorithm and its objective function are chosen for the following reasons. First, it combines value loss and policy loss into a single loss function, which allows a single back-propagation to update all learnable components contributing to this loss function, thus enabling training end-to-end. Second, it has been shown that PPO is sample efficient and works well across a wide range of benchmarks \cite{ppo}. Lastly, PPO can perform roll-outs in vectorised environments and update via mini-batch. This makes distributed training possible, and it is especially important for very large DNNs, where more training is needed to encode the graph attributes. In our empirical evaluation, we find training in a single machine is sufficient, but as model sizes of DNNs continue to grow, distributed training is necessary for the future. We leave accelerating RL training for future work.

Note that our system environment is agnostic to the RL algorithms, therefore it can be reused for different choices. In this work, we focus on the PPO clip variant algorithm.

\subsection{Graph embedding network architectures}
\label{sec:prob:subsec:gnn}

In this section, we provide more details on the GNN architecture, as it is essential for applied RL in the tensor graph superoptimisation domain. The GNN consists of one node update layer, followed by $k$ graph attention layers, and has one final global update layer at the end. This architecture is not designed without careful consideration, as each layer has its responsibility to learn specific representations. The first node update layer uses edge attributes to update node attributes:

    \vspace{-5mm}
\begin{equation}
    \vec{h'}_i = \sigma \{W (\sum_{j\in \mathcal{E}_i} \vec{e}_j \lVert \vec{h_i})\}
    \vspace{-3mm}
\end{equation}

Where $\vec{h'}_i$ is the attribute output of node $i$, given its initial node attribute $\vec{h_i}$ and its incoming edges attributes $\vec{e}_j$. The $\lVert$ operator stands for concatenation as commonly used in GNN terminology, and $\sigma$ is a non-linear activation function. 

This layer is necessary because each node in the graph corresponds to an operator kernel launch in the actual DNN. The kernel launch time is determined by its incoming tensors (edges), and the type of operator, such as MatMul, Conv2D, etc. As a result, the first GNN layer is responsible to learn the kernel launch time of each operator by combining the operator type and the incoming tensor shapes.

The subsequent $k$ graph attention layers (GAT) \cite{gat} are used to learn the topology of the computation graph. This is achieved by message passing between nodes with their neighbouring nodes, and this mechanism has been shown effective in graph-like data than other mechanisms using Euclidean metrics. 

    \vspace{-5mm}
\begin{equation}
    \vec{h'}_i = \sigma (\sum_{j\in \mathcal{N}_i} \alpha_{i, j} W \vec{h}_j) 
    \vspace{-3mm}
\end{equation}

Where $\mathcal{N}_i$ represents neighbours of node $i$, and $\alpha_{i,j}$ is learned during back-propagation. The number of GAT layer $k$ controls how many message-passing steps are performed and is a hyper-parameter. By performing more message-passing steps, a node's message can reach a more distant neighbour but also increases computational demands. The number of message-passing steps is specified in Appendix \ref{appendix:hyper}.

After passing through the GAT layers, each node has updated its representation. The final layer aggregates all nodes' attributes along with the original attribute of the graph to produce a final graph representation:

    \vspace{-5mm}
\begin{equation}
   \vec{g'} = \sigma (\sum_{\mathcal{N}} \vec{h}  \lVert \vec{g})
    \vspace{-3mm}
\end{equation}

The final global update layer is also necessary, as tensor graph superoptimisation changes the tensor graph structure at each iteration. Thus, we need graph-level representation for decision-making. This also distinguishes X-RLflow from other work, such as GO \cite{go1}, where they focus on making node-level decisions like fusion and scheduling. X-RLflow operates on a higher graph-level transformation and therefore is orthogonal to other RL-based optimisers.

\section{Evaluation}
\label{sec:eval}

In this section, X-RLflow is evaluated with TASO and Tensat over a wide range of DNNs. PET is not compared given the discussion in section \ref{sec::limitation}. We seek to answer the following questions:

\begin{enumerate}

\item Can X-RLflow achieve further speedup than the greedy-based method?
\vskip 1mm

\item Can X-RLflow finish optimisation within a tolerable time compared to the existing systems?
\vskip 1mm

\item Once X-RLflow is trained, can it generalise to other tensor shapes given the same computation graph?

\end{enumerate}

\subsection{Experimental setup}

\subsubsection{Platforms}

All experiments presented were performed using a single machine running Ubuntu Linux 18.04 with a 24-core Intel E5-2620@2.00GHz, 256GB RAM and an NVIDIA GeForce GTX 1080. The hardware libraries used for running end-to-end inference are CUDA 10.2 and CuDNN 7.6.5. Each experiment is performed $5$ times and we compute the means and standard deviations. 

\subsubsection{Frameworks}

As previously discussed, we used the open-sourced version of TASO as the graph transformation backend. The RL agent is implemented via \texttt{JAX} \cite{jax2018github} and \texttt{jraph} package \cite{jraph2020github}. While tunning hyper-parameters for RL is notoriously difficult, recent works such as \cite{the_37} perform large-scale studies across a wide range of environments, and suggest empirically good hyper-parameter values. Therefore, we adopt good hyper-parameters from those works, and we keep the values fixed over all experiments. 

\subsubsection{Workloads}

\begin{table}[t]
\caption{Properties of evaluated DNNs. The complexity indicates the average number of candidates at each iteration of the transformation process. Although this number may not comprehensively quantify the search space, a higher number typically indicates there are more combinatorial opportunities throughout the optimisation process.} 
\label{table:eval:graph-props}
\vskip 0.15in
\begin{center}
\begin{small}
\begin{sc}
\begin{tabular}{lccr}
\toprule
DNNs & Type & Complexity  \\
\midrule
         InceptionV3 & Convolutional & 50 \\
         Squeezenet & Convolutional & 20 \\
         ResNext-50 & Convolutional & 13 \\
         BERT & Transformer & 26 \\
         DALL-E & Transformer  & 20 \\
         T-T & Transformer & 25 \\
         ViT & Transformer & 32 \\
\bottomrule
\end{tabular}
\end{sc}
\end{small}
\end{center}
\vskip -0.1in
\end{table}

We have chosen state-of-the-art DNNs over a wide range of tasks in our evaluation. InceptionV3 \cite{szegedy2015rethinking} is a common, high-accuracy model for image classification trained on the ImageNet dataset. ResNext-50 \cite{he2015deep} is also a deep convolutional network for vision tasks. SqueezeNet \cite{iandola2016squeezenet} is a shallower yet accurate model on the same ImageNet dataset. BERT \cite{devlin2019bert}, ViT \cite{nayak2019}, DALL-E \cite{dalle}, T-T \cite{tt}, are large transformer networks that succeed across a wide range of tasks, including visions, languages, texts and audios. Although the common building block is multi-head attention, they are often combined with different neural network blocks and thus overall have diverse computation graph structures. Figure \ref{table:eval:graph-props} lists the properties of those DNNs. Note that some tensor operators are not supported by TASO and we simply skip those operators when building the computation graph.

\subsection{End-to-end speedup}
\label{sec:eval:subsec:mb:sec:runtimeperf}

\begin{figure}
  \centering
  \includegraphics[width=1.0\columnwidth]{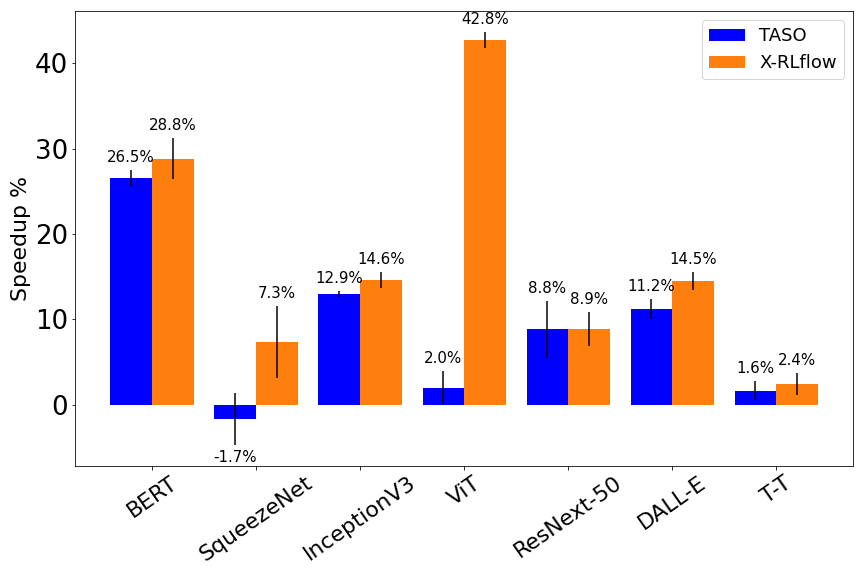}
  \caption{End-to-end inference speedup by TASO and X-RLflow. In each case, the evaluation is run five times to measure the mean and standard deviations.}
  \label{fig:eval:e2e}
\end{figure}

Figure \ref{fig:eval:e2e} shows the end-to-end inference latency speedup of TASO and X-RLflow. For TASO, we run with its default setting as in its artefact evaluation. For X-RLflow, each agent was trained from the respective graph, as described in Section\ref{sec::optnns}, and they are trained for over $1000$ episodes. All agents use the same reward function as Equation \ref{eq. reward func} as well as the same set of hyper-parameter, and common hyper-parameters are listed in Appendix \ref{appendix:hyper}.

Firstly, we can observe among all cases, X-RLflow achieves better speedup than TASO's substitution engine. This verifies that X-RLflow can learn to make better decisions for long-term rewards and leverage the end-to-end feedback signal to guide its decision-making. X-RLflow finds more globally optimal tensor graph structures after the transformation process. 

We can also observe there are two DNNs where X-RLflow achieves much better speedup than TASO. In the case of SqueezeNet, TASO achieves a negative speedup. This is because the cost model is inaccurate, and it misleads the substitution engine. Note that the cost modelling depends on the execution hardware, so different GPUs may be evaluated differently. The cost modelling issue is also reported by Tensat, where it affects the extraction of the E-graph. 

For ViT, we observe an over $40\%$ speedup. While this appears to be a special case, we observe that after a sequence of graph-level optimisation, some operators have no data dependencies so they can be pre-processed before actually running inference. This is similar to constant folding in compiler optimisation. Cost modelling does not consider constant folding because it simply sums over all operators' execution times. Thus, by using the end-to-end feedback signal, X-RLflow can discover better tensor graph structures by considering downstream optimisation opportunities. This also indicates more optimisation opportunities can be revealed by combining the compiler optimisation pipeline across different layers of IRs.

\subsection{Transformation heatmap}
\label{sec:Transformation heatmap}

\begin{figure}
  \centering
  \includegraphics[width=1.0\columnwidth]{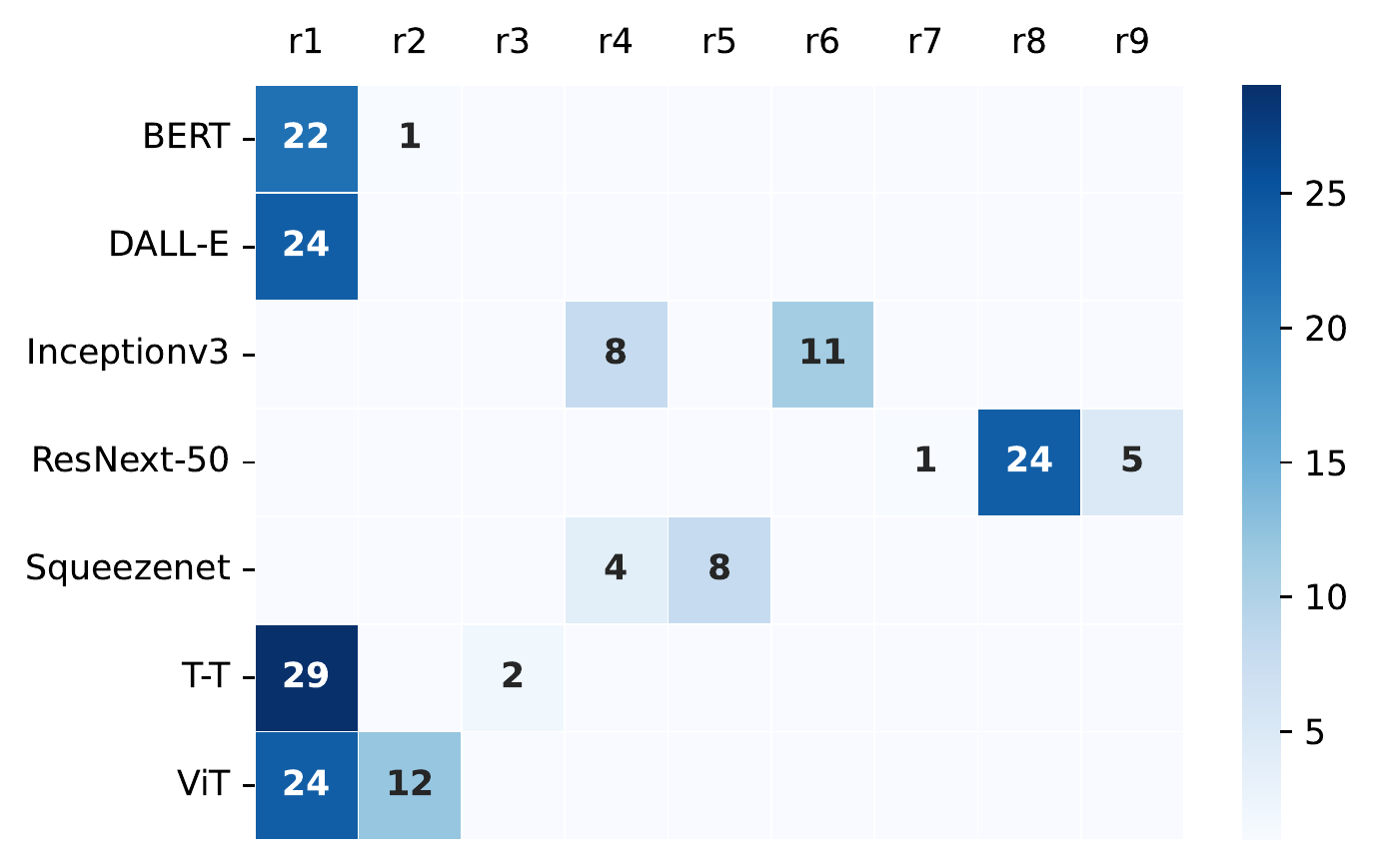}
  \caption{This heatmap shows the rewrite rules applied by X-RLflow. Although there are over 100 possible rewrite rules, we only show the rules applied at least once. The count for each rewrite rule shows the number of times it has been applied. Note that each application of the rewrite rule takes place at different locations of the target computation graph. A higher count means that X-RLflow can find a long sequence of subgraph transformation before termination. This can indicate X-RLflow has discovered a performant graph via the sequence of subgraph transformation.} 
  \label{fig:eval:xfer-heatmap}
\end{figure}

Figure \ref{fig:eval:xfer-heatmap} shows the counts of rewrite rules that are applied to the tensor graphs during the optimisation process. Notably, DNNs that primarily consist of convolutional operators and the ones that consist of matrix multiplication are targeted by different tensor graph rewrite rules. In general, convolutional operators can be rewritten by more rules, but they have shorter transformation sequences, meaning they are less beneficial by long-term delayed reward. Transformer-type of DNNs on the other hand, are targeted by fewer rewrite rules but have longer transformation sequences. We conclude that RL by design maximises long-term rewards, and thus it shows more advantages in long sequential tasks.

\subsection{Optimisation time}

\begin{figure}
  \centering
  \includegraphics[width=1.0\columnwidth]{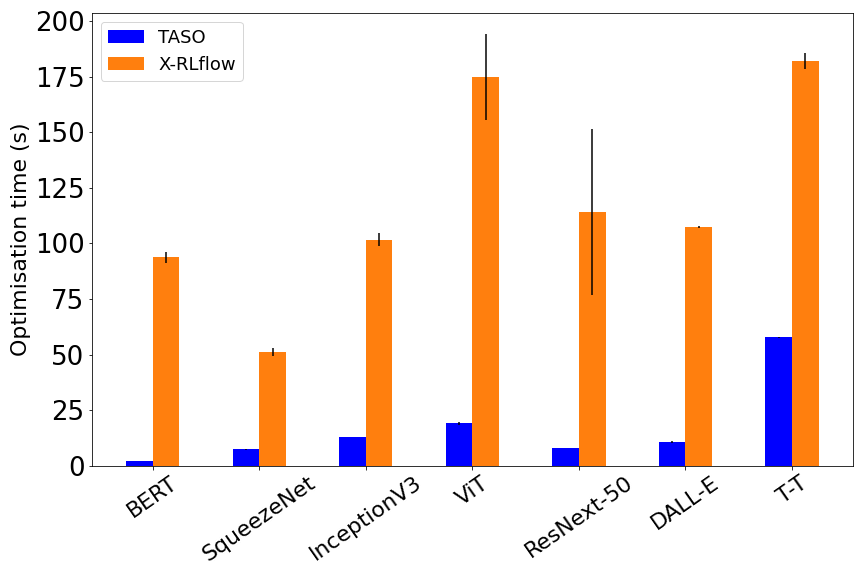}
  \caption{Optimisation time taken for TASO and X-RLflow.}
  \label{fig:eval:optimisation-time}
\end{figure}

Figure \ref{fig:eval:optimisation-time} shows the optimisation time required to transform the optimised graph of the evaluated DNNs. We note that the optimisation time for the X-RLflow does not include the time needed to train the RL agents. We can see TASO in all cases take less than $75$ seconds to generate its optimised graph given its default setting. Although TASO has a configurable optimisation time budget, increasing the budget does not give more speedup while causing a much longer optimisation time. For example, the optimisation time is $10$x more while the performance increase is less than $5\%$. Therefore, We conclude this phenomenon indicates greedy substitution is stuck in the local optimum and we use the default budget for TASO.

X-RLflow generally takes more time to optimise, because the RL agent performs a forward pass to make decisions at every iteration of the transformation. This can be accelerated by putting the agent's networks in GPU. However we only have one GPU and its memory is pre-allocated by the backend of the environment, i.e. TASO, so we are unable to do that. The optimisation time can be reduced significantly if the agent's inference can be accelerated by another GPU. Even in the case of the CPU, we can see it takes less than $200$ seconds in optimisation, which is affordable before model deployment.

\subsection{Generalisation to different tensor shapes}

\begin{figure}
  \centering
  \includegraphics[width=1.0\columnwidth]{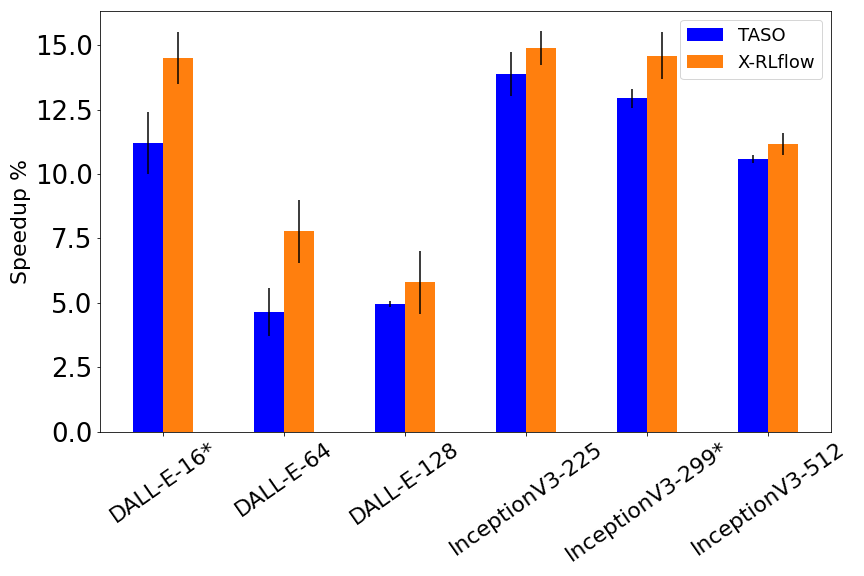}
  \caption{Generalisation to different tensor shapes on DALL-E and InceptionV3. The suffix number following the name of DNNs indicates the input tensor shape. For example, `InceptionV3-225' indicates the input image has a height and width of $225$. `*' indicates the DNNs where X-RLflow is trained to optimise.} 
  \label{fig:eval:var-shape}
\end{figure}

Figure \ref{fig:eval:var-shape} shows the generalisation ability of X-RLflow to various tensor shapes on DALL-E and InceptionV3 respectively. This is possible when an update in the data processing pipeline results in different tensor shapes to input to DNN models. We show that by training X-RLflow in a static shape environment once, it is sufficient to generalise and achieve good performance for different input tensor shapes. 

Performing training on multiple graphs and generalising to unseen DNNs is a more significant generalisation direction we would like to investigate. However, this poses a challenge to the GNN encoder, as it is responsible to embed diverse computation graphs. As such, it is much more computationally expensive to achieve held-out graph generation. We will leave this for future work. 

\subsection{Comparison with Tensat}
\label{sec:eval:subsubsec:tensat}

\begin{figure}
  \begin{center}
    \includegraphics[width=1.0\columnwidth]{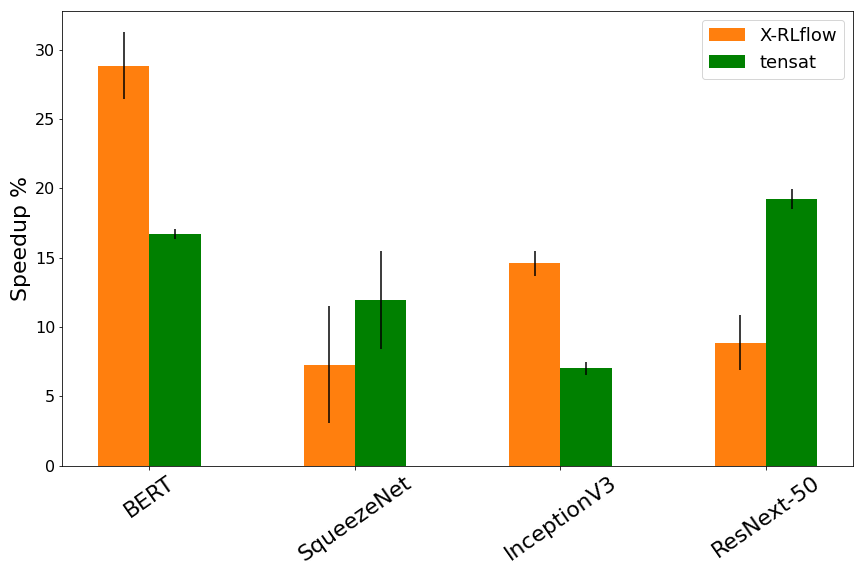}
  \end{center}
  \caption{Comparison of end-to-end DNNs speedup with Tensat.} 
  \label{fig:eval:with-tensat}
\end{figure}

Tensat is another tensor graph superoptimiser that employs equality saturation as mentioned in section \ref{sec::Tensor graph structure superoptimisation systems}. Figure \ref{fig:eval:with-tensat} shows a comparison of X-RLflow with Tensat.

We can see that X-RLflow outperforms Tensat in BERT and InceptionV3, but falls short in Squeezenet and ResNext-50. Note that Tensat is sensitive to its parameters, and we use the default values as reported in their artefact evaluation. By varying Tensat's parameters, it is possible to have a performance increase but also raises the risk of running out of memory and optimisation timeout.

By examining the dataflow graph of SqueezeNet, InceptionV3 and ResNext-50, we notice that while they all primarily consist of convolution blocks, InceptionV3 has more unique blocks in its computation graph, indicating more combinatorial optimisation opportunities. This is also verified by the ``complexity'' as shown by Table \ref{table:eval:graph-props}. As RL favours more combinatorial chances, X-RLflow is more likely to perform well in complex graphs. 

X-RLflow outperforms Tensat when optimising BERT. This is because BERT's multi-head attention blocks mainly consist of matrix multiplication, and there is a specific ``multi-pattern rewrite rule'' for matrix multiplication in Tensat that can grow the E-graph extremely large. As a result, Tensat has to limit the application of multi-pattern rewrite rules up to a constant $k$. In their default setting, $k$ is set to $1$, and this is certainly not enough to explore all rewrite opportunities for BERT. Increasing $k$ will likely increase performance, but also dramatically increase the optimisation time and memory usage. To know more about the multi-pattern rewrite rules, we refer interested readers to the section $4$ of Tensat's paper. X-RLflow does not use the E-graph data structures, so it does not suffer from the multi-pattern rewrite rule issues. 

We also want to run more experiments on transformer types of DNNs, but we are unable to do so, because Tensat needs to convert TASO's graph representation to its S-exprs representation, and strictly filters out cycles when building the E-graph. When we try to add a new transformer, the cycle filtering algorithm reports an error. As such, we are unable to perform more experiments except for those provided by Tensat. On the contrary, X-RLflow does not use a new representation, and therefore it can optimise any computation graph as long as it is supported by TASO. 

In future work, we would like to combine X-RLflow with Tensat. Specifically, as the E-graph is not saturated, Tensat re-introduces the phase-ordering issue when building the E-graph, which can be addressed by RL. On the other hand, the E-graph can compactly represent many graph IRs, which can decrease the state space for RL. We believe combining RL with equality saturation will lead to better performance for optimising computation graphs.

\section{Related Work}

\subsection{Optimisation of computation graphs}

Rule-based approaches such as those used in TensorFlow \cite{tensorflow2015whitepaper, 199317} and TVM \cite{chen2018tvm} use a pre-defined set of transformations to optimise computation graphs. On the contrary, tensor graph superoptimisers, such as \cite{jia2019optimizing,jia2019taso,tensat,wang2021pet}, automatically search for transformations to apply to the input graph. In addition, OCGGS \cite{graph_opt} theoretically proves the graph substitutions problem to be NP-hard, and comes up with a dynamic programming approach to exactly solve the problem. However, the exact solution is impractical due to the long search time. As a result, an approximate sampler is also proposed to solve this problem. However, it fails to achieve better performance than existing systems. We have described in detail in section \ref{sec::Tensor graph structure superoptimisation systems} the advantages and drawbacks of each system, and our motivation for using RL to tackle this problem. We show that the optimisation process is more globally optimal and generalisable with X-RLflow. 

\subsection{RL in system optimisation}

There has been an effort to apply reinforcement learning to system optimisation. Due to its theoretical soundness for handling sequential decision-making problems, RL has been applied to circuits design \cite{prefix_rl}, the compiler passes ordering \cite{autophase}, and datacenter control \cite{dc_control}. Among existing works, NeuRewriter \cite{Neurorewriter} is a particularly relevant work that implements an RL-based rewrite system to perform expression rewrite, job scheduling and vehicle routing. While related, X-RLflow performs optimisation in the tensor graph domain with a specific design of the RL agent and the environment to handle tensor graph transformation. As such, it is complementary to NeuRewriter. 

GO \cite{go-1} is an RL-based graph optimiser that performs node fusion, node scheduling and device placement at once. As a result, GO outputs decisions for each node in the computation graph. On the other hand, X-RLflow is specific to graph-level optimisation, so the tensor graph structure changes dynamically throughout the optimisation process. As such, X-RLflow is orthogonal to existing works and can be used to optimise before performing downstream optimisation, such as GO.

\subsection{Model-based reinforcement learning}

Model-based RL is a class of reinforcement learning algorithms in which we aim to learn a model (or use a given model) of the real environment where an agent acts. The world model \cite{ha2018worldmodels} proposed to learn the environment dynamics using recurrent neural networks. Alternative approaches have also been proposed such as imagination-augmented agents \cite{weber2018imaginationaugmented} and model-based value estimation for model-free agents \cite{feinberg2018modelbased}. The main benefit of model-based reinforcement learning is a world model help reduce the agent interaction with the environment, thus accelerating the training process. Moreover, the world model may provide better generalisation ability because it can model the latent transition of the state space. Recent advances in model-based reinforcement learning show a world model agent can outperform its model-free counterpart across diverse domains including vision and control tasks, with a fixed set of hyper-parameters. X-RLflow may be combined with those methods to have better sample efficiency and generalisation ability because the environment transition does not assume a fixed RL algorithm. 


\section{Limitations and future work}

There are a few limitations with X-RLflow. First, training RL is computationally expensive and time-consuming. Potential solutions to mitigate the problem includes setting up a distributed training environment, where training data can be generated in parallel. This allows trading off training time with computing power. On the other hand, we could exploit the layer structure of DNNs, and simply perform optimisation on sub-graphs individually. However, this requires certain heuristics to avoid missing out on optimisation opportunities across different sub-graphs. Second, X-RLflow cannot be generalised to unseen tensor graphs at the moment. We expect by adopting methods from model-based RL and training the agent across various DNNs, X-RLflow can gain stronger generalisation ability. Alternatively, combining RL with equality saturation can reduce the state space of RL because E-graphs can compactly represent many graph IRs, and this may be another direction to explore.

\section{Conclusion}

In this work, we present X-RLflow as a novel end-to-end tensor graph structure superoptimiser. We explain our motivations for using RL and describe our formulation of the RL algorithm in the tensor graph domain. We also present the architecture design of X-RLflow in detail. Various experiment results show that X-RLflow can achieve better performance over a wide range of evaluation DNNs and have up to $40\%$ speedup over state-of-the-art systems. We also demonstrate its generalisation ability by performing inference in unseen environments. We argue that the applicability of RL in a delayed reward environment sheds light on system optimisation when the feedback signal is expensive. The effectiveness of X-RLflow suggests that it is a promising direction for tensor graph structure superoptimisation.

\section*{Acknowledgements}

We thank Sam Ainsworth, Amitabha Roy, Wenjun Hu,  Sami Alabed for their valuable comments and feedback. We also thank Zhihao Jia for insightful discussions at an early stage of our work.

\bibliography{main}
\bibliographystyle{mlsys2023}


\onecolumn
\newpage
\appendix
\section*{Appendices}
\addcontentsline{toc}{section}{Appendices}
\renewcommand{\thesubsection}{\Alph{subsection}}

\subsection{Hyper-parameters}
\label{appendix:hyper}

\begin{table}[H]
  \caption{Hyper-parameter values}
  \begin{center}
    \begin{tabular}[H]{ccc}
      \toprule
      Name         & Value          & Explanation  \\ \midrule
      Learning rate   & $5e-4$ & Learning rate of PPO's policy and value networks   \\
      Value loss coefficient($c_1$) & $0.5$ &  Value loss coefficient  \\
      Entropy loss coefficient($c_2$) & $0.01$ &  Entropy loss coefficient  \\
      Edge attribute constant (M)   & $4096$ & The constant to normalise all edge attributes   \\
      Number of GAT layers (k)   & $5$ & Number of GAT layers   \\
      Update frequency   & $10$ & The frequency to perform an update   \\
      feedback frequency (N)   & $5$ & The frequency to perform an end-to-end inference and return as reward   \\
      MLP heads   & $[256, 64]$ & The MLP hidden neurons for policy and value networks   \\
      Batch size      & 16   & Batch size for updating agent's networks   \\ \bottomrule
  \end{tabular}
  \end{center}
  \label{tab:hyper-parameters}
\end{table}



\end{document}